\documentclass{article}
\pdfoutput=1

\usepackage[numbers]{natbib}
\usepackage{amsmath}
\usepackage{amsfonts}
\usepackage{amssymb}
\usepackage{xcolor}
\usepackage{authblk}

\newcommand{\X}{\mathcal{X}}
\newcommand{\Y}{\mathcal{Y}}
\newcommand{\F}{\mathcal{F}}
\newcommand{\reals}{\mathbb{R}}
\newcommand{\PP}{\mathbb{P}}
\newcommand{\PPP}{\text{PP}}
\newcommand{\E}{\mathbb{E}}
\newcommand{\BeP}{\text{BeP}}
\newcommand{\T}{\mathcal{T}}

\newcommand{\DP}{\text{DP}}

\usepackage{comment}

\title{A survey of non-exchangeable priors for Bayesian nonparametric models}
\author[1]{Nicholas~J.~Foti}
\author[2]{Sinead~Williamson}
\affil[1]{Department of Computer Science, Dartmouth College}
\affil[1]{Machine Learning Department, Carnegie Mellon University}

\begin{document}


\maketitle

\begin{abstract}

Dependent nonparametric processes extend distributions over measures, such as
the Dirichlet process and the beta process, to give distributions over
collections of measures, typically indexed by values in some covariate space.
Such models are appropriate priors when exchangeability assumptions do not
hold, and instead we want our model to vary fluidly with some set of
covariates. Since the concept of dependent nonparametric processes was formalized by
MacEachern \cite{MacEachern:2000}, there have been a number of models
proposed and used in the statistics and machine learning literatures. Many of
these models exhibit underlying similarities, an understanding of which, we
hope, will help in selecting an appropriate prior, developing new models, and
leveraging inference techniques.  
\end{abstract}

\section{Introduction} 

There has recently been a spate of
papers in the statistics and machine learning literature developing dependent
stochastic processes  and using them as priors in Bayesian nonparametric models.  In this
paper, we aim to provide a representative snapshot of the currently available
models, to elucidate links between these models, and to provide an orienting
view of the modern constructions of these processes.  

Traditional nonparametric priors such as the Dirichlet process
 \citep[DP,][]{Ferguson:1973}, Chinese
restaurant process \citep[CRP,][]{Pitman:2002}, Pitman-Yor process
\cite{Perman:Pitman:Yor:1992} and
the Indian buffet process \citep[IBP,][]{Griffiths:Ghahramani:2005} assume that 
our observations are exchangeable.  Under the assumption of exchangeability 
the order of the data points does not change the probability distribution.  

Exchangeability is not a valid assumption for all data. For example,
in time series and spatial data, we often see correlations between
observations at proximal times and locations. The fields of time series
analysis \cite{Tsay:2010} and spatial statistics \cite{Cressie:2011} exist to model
this dependence.  In fact, many data sets contain covariate information, that is variables
we do not wish to model but only condition on, that we may desire
to leverage to improve model performance.  

The use of non-exchangeable priors in a Bayesian nonparametric context
is relatively new. While not the first model to address
non-exchangeability in a nonparametric framework (this honor arguably
goes to \cite{Cifarelli:Regazzini:1978}), the seminal work in this area is
a technical report by MacEachern \cite{MacEachern:2000}.
In that paper, MacEachern formally introduces the idea of dependent
nonparametric processes, and proposes a set of general desiderata, described in Section~\ref{sec:DDPintro}, that paved
the way for subsequent work. 

Loosely, dependent nonparametric
processes extend existing nonparametric priors over measures,
partitions, sequences, etc. to obtain priors over collections of
such structures. Typically, it is assumed that the members of these
collections are associated with values in some metric covariate space, such
as time or geographical location, and that locations that are close in
covariate space tend to generate similar structures. 



Since MacEachern's technical report, there has been an explosion in the number of such
dependent nonparametric processes. While, at first glance, the range
of models can seem overwhelming, the majority of existing models fall
under one (or sometimes more) of a relatively small number of classes:

\begin{itemize}
  \item Hierarchical models for grouped data with categorical covariates.
  \item Collections of random measures where the atom locations vary
    across covariate space.
  \item Collections of random measures where dependency is introduced
    via a stick-breaking representation.
  \item Collections of locally exchangeable sequences generated by inducing
    dependency in the family of conditional distributions in a de
    Finetti representation.
  \item Collections of locally exchangeable sequences obtained my
    modifying the predictive distribution of an exchangeable
    sequence. 
  \item Collections of random measures obtained by exploiting
    properties of the Poisson process.
  \item Collections of random measures obtained by superpositions and
    convex combinations of random measures.
\end{itemize}

We find there are many similarities
between the models in each class. There are also differences, particularly in
the form of covariate space assumed:  Some models are appropriate only
for time
\cite{Lin:Grimson:Fisher:2010,Caron:Davy:Doucet:2007,Griffin:2011},
while others consider general underlying covariates 
\cite{Williamson:2012,Williamson:Orbanz:Ghahramani:2010,Zhou:Yang:Sapiro:Dunson:Carin:2011,Ren:Wang:Dunson:Carin:2011}. Different
authors have applied varying inference techniques to highly related
models, and have adapted their models to numerous applications.

By pointing out similarities between existing models, we hope
to aid understanding of the current range of models, to assist in
development of new models, and to elucidate areas where inference
techniques and representations can easily be transferred from one model
to another. 

This is not the first survey of dependent nonparametric processes: in
particular, a well-written survey of some of the earlier constructions
can be found in \cite{Dunson:2010}. However, the subfield is growing
rapidly, and Dunson's survey does not incorporate much recent work. In particular,
while early research in this area focused on dependent Dirichlet
processes, the machine learning community in particular has recently
begun exploring alternative dependent stochastic processes, such as
dependent beta processes
\cite{Williamson:2012,Zhou:Yang:Sapiro:Dunson:Carin:2011,Ren:Wang:Dunson:Carin:2011,Saeedi:2011}. 

In addition to describing the range of dependent nonparametric priors in the current literature, we
explore some of the myriad applications of these models in Section~\ref{sec:applications}. 

The reader will realize, while reading this paper, that not all of the
models presented fit into MacEachern's original specification. For
example, many of the kernel-based methods do not exhibit easily
recognizable marginal distributions, and some of the models do not
satisfy marginal invariance. In our conclusion, we discuss how the
original desiderata of \cite{MacEachern:2000} pertains to the current body of 
work, and consider the challenges currently facing the subfield.

\section{Background}\label{sec:bg}

The dependent nonparametric models discussed in this survey are
based on a relatively small number of stationary Bayesian
nonparametric models, which we review in this section. We focus on
nonparametric priors on the space of $\sigma$-finite measures. In particular, we consider two classes of random
measure: random probability measures, and completely random
measures. We also discuss exchangeable sequences that are related to
these random measures.
\subsection{Notation}
We introduce some notation here for convenience.  We denote unnormalized random
measures as $G$ and random probability measures as $P$.  We use $\X$ for an
arbitrary covariate space with elements $x,x',\mu \in \X$.  When the covariate
is time, $\X = \reals_+$, we use $t$ for an observed covariate.  We denote a
random measure evaluated at a covariate value $x\in\X$ as $G^{(x)}$, and for an indexed
set of covariates, $\{x_i\}$, as $G_i \equiv G^{(x_i)}$.  The notation $G_i$ is
used extensively when we observe uniformly spaced observations in time.

\subsection{Completely random measures}\label{sec:CRM}
A completely random measure \citep[CRM,][]{Kingman:1967} is a distribution over measures on some measurable space $(\Theta, \F_\Theta)$, such
that the masses $G(A_1), G(A_2), \ldots$ assigned to disjoint subsets
$A_1, A_2, \ldots \in \F_\Theta$ by a draw $G$ from the CRM are
independent random variables. 

A CRM on some space $\Theta$ is characterized by a positive L\'{e}vy measure
$\nu(d\theta, d\pi)$ on the product space $\Theta \times \reals_+$, with
associated product $\sigma$-algebra $\F_\Theta \otimes \F_{\reals_+}$.
Completely random measures have a useful representation in terms of Poisson
processes on this product space.  Let $\Pi = \{(\theta_k,
\pi_k) \in \Theta \times \reals_+\}_{k=1}^\infty$ be a Poisson process on 
$\Theta \times \reals_+$ with
rate measure\footnote{The rate measure is $\E[\Pi(A)]$, that is the expected
number of points of the Poisson process that fall in a measurable set $A$.} given 
by the L\'{e}vy measure $\nu(d\theta,d\pi)$.  We denote this as $\Pi \sim
\PPP(\nu)$.  Then the
completely random measure with L\'{e}vy measure $\nu(d\theta,d\pi)$ can be
represented as 
\begin{equation}
  G = \sum_{k=1}^\infty \pi_k \delta_{\theta_k}
\end{equation}
(see \cite{Kingman:1967} for details).  We denote a draw from a CRM with L\'{e}vy
measure $\nu$ as $G \sim \text{CRM}(\nu)$. Due to the correspondence between CRMs and Poisson processes, we can
simulate a CRM by simulating an inhomogeneous Poisson process with
appropriate rate measure.  

An alternative representation of CRMs, often
referred to as the Ferguson-Klass representation and more generally
applicable to pure-jump L\'{e}vy processes, is given by
transformation of a unit-rate Poisson process on $\reals_+$ \cite{FK:1972}.   Specifically, let
$u_1,u_2,\ldots$ be the arrival times of a unit-rate Poisson process
on $\reals_+$, and denote the tail $T(s)$ of the L\'{e}vy measure
$\nu$ as $T(s) = \nu(\Omega, (s,\infty))$. Then the strictly
decreasing atom sizes of a CRM are obtained by transforming the
strictly increasing arrival times $u_1<u_2<\dots$ according to
$\pi_k = T^{-1}(u_k)$. 

If the L\'{e}vy measure is homogeneous -- ie $\nu(d\theta, d\pi) =
\nu_\theta(d\theta)\nu_\pi(d\pi)$ -- the atom locations are
distributed according to $\nu_\theta (d\theta)$. If the L\'{e}vy
measure is inhomogeneous, we can obtain the conditional distribution
of an atom's location given its size by solving the transport problem
described in \cite{Orbanz:Williamson:2011}.


The class of completely random
measure includes important distributions such as the beta process,
gamma process, Poisson process, and the stable subordinator. Such
distributions are often used as priors on the hazard function in
survival analysis. See \cite{Lijoi:Pruenster:2010} for a recent review
of completely random measures and their applications.

\subsection{Normalized random measures}\label{sec:NRM}
Since any $\sigma$-finite measure on $\Theta$ implies a probability
measure on $\Theta$, we can construct a distribution over probability measures
by normalizing the output of a completely random measure. The resulting class
of distributions are referred to as normalized random measures
\citep[NRM,][]{Lijoi:Pruenster:Regazzini:2003}. Specifically, for $G =
\sum_k \pi_k\delta_{\theta_k} \sim
\text{CRM}(\nu)$, define the probability measure
\begin{equation}
    \label{<++>}
    P = \sum_{k=1}^\infty \frac{\pi_k}{\sum_{j=1}^\infty \pi_j}
        \delta_{\theta_k} \sim \text{NRM}(\nu)\, .
\end{equation}

The most commonly used exemplar is the Dirichlet process, which can be obtained as a normalized gamma process. Distributions over probability measures are of great importance in Bayesian statistics and machine learning, and normalized random measures have been used in
many applications including natural language
processing , image segmentation and speaker diarization \cite{Teh:Jor:2010a}.
\subsection{Exchangeable sequences}
Often, we do not work directly with the random measures described in
Sections~\ref{sec:CRM} and \ref{sec:NRM}. Instead, we work with closely related exchangeable nonparametric sequences. Recall that an infinitely exchangeable sequence is one whose probability is invariant under finite permutations $\tau_n$ of the first $n$ elements, for all $n\in\mathbb{N}$ \cite{Aldous:1983}.
De Finetti's theorem tells us that any infinitely exchangeble sequence $X_1,X_2,\dots$ can be written as a mixture of i.i.d. samples
\begin{equation}
  \label{eqn:definetti}
  \PP(X_1,X_2,\ldots,X_n) = \int \prod_{i=1}^n Q_\theta(X_i) P(d\theta)
\end{equation}
where $\{Q_\theta , \theta \in \Theta\}$ is a family of conditional distributions and $P$ is a distribution over $\Theta$ 
called the de Finetti mixing measure. 

We can obtain a number of interesting exchangeable sequences by letting $\Theta$ be a space of measures and $P$ be a distribution over such measures. For example, if $P$ is a Dirichlet process, and $Q_\theta$ is the discrete probability distribution described by the probability measure $\theta$, then we obtain a distribution over exchangeable partitions known as the Chinese restaurant process \citep[CRP,][]{Aldous:1983}.

Similarly, we can use a completely random measure as the de Finetti
mixing measure, to obtain a sequence of exchangeable vectors (or
equivalently, a matrix with exchangeable rows). For example, if we
combine a beta process prior with a Bernoulli process likelihood, and
integrate out the beta process-distributed random measure, we obtain a
distribution over exchangeable binary matrices referred to as the
Indian buffet process  \citep[IBP,][]{Griffiths:Ghahramani:2005,
  Thibaux:Jordan:2007}. The IBP has been used to select subsets of an infinite set of latent variables in nonparametric versions of latent variable models such as factor analysis and independent component analysis \cite{Knowles:Ghahramani:2007}. Other distributions over
exchangeable matrices have been defined using the beta process
\cite{Zhou:Hannah:Dunson:Carin:2012} and the gamma process \cite{Saeedi:2011,Titsias:2007} as mixing
measures. 

Conjugacy between the de Finetti mixing measure and the family of conditional
distributions often means the predictive distribution $P(X_{n+1}|X_1,\dots,
X_n)$ can be obtained analytically. For example, the predictive distribution
for the Chinese restaurant process  is given by
\begin{equation}
    \label{eqn:crp}
    p(X_{n+1}|X_1,\dots,X_n) = \begin{cases}\frac{m_k}{n+\gamma} &\mbox{if }m_k>0\\
    \frac{\gamma}{n+\gamma} &\mbox{otherwise}
    \end{cases}
\end{equation}
where $m_k$ is the number of observations in $X_1,\dots, X_n$ that have been 
assigned to cluster $k$, and $\gamma$ is the concentration parameter of the 
underlying Dirichlet process.
The distribution represented by Equation~\ref{eqn:crp} can be
understood in terms of the following analogy. Consider a sequence of
customers entering a restaurant with an infinite number of tables,
each serving a single dish. Let $z_n$ denote the table that
customer $n$ sits at, and let $K^+$ denote the number of tables occupied by at
least one customer.
Customer $n+1$ enters the restaurant and sits at table $k$ where dish $\theta_k$ 
is being served, with probability proportional to $m_k$, the number of previous customers 
to have sat at table $k$. This is indicated by setting $z_{n+1}=k$. 
Alternatively, the customer may sit at a new table, $z_{n+1}= K^++1$, with 
probability proportional to $\gamma$ and chooses a new dish, $\theta \sim H_0$, for 
some measure $H_0$.

\section{Dependent nonparametric processes}\label{sec:DDPintro}
If we define a nonparametric process as a distribution over measures with
countably infinite support, then a \emph{dependent} nonparametric process is a
distribution over \emph{collections} of such measures. Joint distributions over
discrete measures have been proposed since the early days of Bayesian
nonparametrics (for example \cite{Cifarelli:Regazzini:1978}), but a formal
framework was first proposed in a technical report of MacEachern
\cite{MacEachern:2000}. MacEachern proposed the following criteria for a
distribution over collections $\{G^{(x)}, x \in \X\}$ of measures:
\begin{enumerate}
  \item The support of the prior on $\{G^{(x)} : x \in \Lambda\}$ for
    any distinct set $\Lambda$ should be large.
  \item The posterior distribution should be reasonably easy to obtain, either
    analytically or computationally.
  \item The marginal distribution of $G^{(x)}$ should follow a familiar
    distribution for each $x \in \X$.
  \item If a sequence of observations converges to some $x_0$ then the
    posterior distribution of $G^{(x)}$ will also converge to some 
    $G^{(x)}_0$.
\end{enumerate}

This specification is vague about the form of the dependence and of the space
$\X$. It is typically assumed that $(\X, d)$ is some metric space, and that as
$x^\prime \rightarrow x$ then $G^{(x^\prime)}\rightarrow G^{(x)}$.  There are, however, distributions in the class of dependent nonparametric processes that do not require a metric space. Models such as the hierarchical Dirichlet process \cite{Teh:Jordan:Beal:Blei:2006} create distributions over exchangeable, but not independent, measures. Other models that fall under this category include the hierarchical beta process \cite{Thibaux:Jordan:2007} and a number of partially exchangeable models \cite{Cifarelli:Regazzini:1978, Muller:Quintana:Rosner:2004}.

In this survey, we focus on distributions over
collections of measures indexed by locations in some metric
space. Examples of typically used spaces include $\mathbb{R}_+$ (for
example, to represent continuous time), $\mathbb{N}$ (for example, to
represent discrete time), or $\mathbb{R}^d$ (for example, to represent
geographic location). We will, however, spend a little time discussing
the two most commonly used non-covariate-indexed dependent
nonparametric processes, the hierarchical Dirichlet process and the
hierarchical beta process, since they are frequently used as part of
other dependent nonparametric processes.

We will also consider distributions over collections of partitions and vectors
that can be seen as dependent extensions of models such as the CRP and
the IBP. It is clear, for example, that any dependent Dirichlet
process (DDP) can be used to
construct a dependent CRP, by constructing a distribution over
partitions using the marginal measure at each covariate
location. However, we will see that we can also construct dependent
Chinese restaurant processes and Indian buffet processes by directly
manipulating the corresponding stationary model, and that such
distributions do not necessarily correspond to a simple random measure interpretation.


\section{Distributions over exchangeable random measures}

While most of the processes we will present in this paper
consider collections of measures indexed by locations in some
covariate space endowed with a notion of similarity, the 
definition provided by MacEachern does not
require such a space. In fact, one of the most commonly used dependent
nonparametric processes is a distribution over an exchangeable
collection of random measures.

The hierarchical Dirichlet process
\citep[HDP,][]{Teh:Jordan:Beal:Blei:2006} is a distribution over
multiple correlated discrete probability measures with a shared
support (that is, the locations of the atoms are shared across the
random measures. 
Each probability measure is distributed according to a
Dirichlet process, with a shared concentration parameter and base
measure. In order to ensure sharing of atoms, this base measure must
be discrete; this is achieved by putting a Dirichlet process prior on
the base measure, resulting in the following hierarchical model:
\begin{equation}
\begin{split}
G_0 &\sim \DP(\gamma_0, H) \\
G_j &\sim \DP(\gamma, G_0) \, , \, j=1,2,\dots
\end{split}
\end{equation}

The HDP is particularly useful in admixture models, where each data
point is represented using a mixture model, and components of the
mixture model are shared between data points. It has been used in a
number of applications including text modeling \cite{Teh:Jordan:Beal:Blei:2006}, 
infinite hidden Markov models \cite{Teh:Jordan:Beal:Blei:2006,Fox:2008}, 
and modeling genetic variation between and within populations
\cite{Xing:Soh:Jor:Teh:2006}. Compared with most of the covariate-dependent
nonparametric processes described in this survey, inference in the HDP
is relatively painless; a number of Gibbs samplers are proposed in
\cite{Teh:Jordan:Beal:Blei:2006}, and several variational approaches
have also been used
\cite{Teh:Kur:Wel:2008,Wang:Paisley:Blei:2011,Bryant:Sudderth:2012}.
Additionally, sequential Monte Carlo methods \cite{Doucet:2001} have been 
developed for HDP topic models \cite{Ahmed:2011}.

Other hierarchical nonparametric models can be constructed in a
similar manner. The hierarchical beta process
\cite{Thibaux:Jordan:2007} replaces the hierarchy of Dirichlet
processes with a hierarchy of beta processes, allowing the creation of
correlated latent variable models.

As we will see throughout this paper, hierarchical models such as
these are often used in conjunction with an explicitly
covariate-dependent random measure, for example by replacing the
DP-distributed base measure $G_0$ with a DDP-distributed collection of
base measures.


\section{Dependence in atom location}\label{sec:var_atoms}

Consider a distribution over collections of atomic measures
\begin{equation*}
    \phantom{,}\{G^{(x)}:=\sum_{k=1}^\infty \pi_k^{(x)}\delta_{\theta_k^{(x)}}, x \in \X\}.
\end{equation*}
One of the simplest ways of inducing dependency is to assume a shared set of
atom sizes $\pi_k^{(x)} = \pi_k, k=1,2,\dots,  x \in \X$, and allowing the
corresponding atom locations $\theta_k^{(x)}$ to vary according to
some stochastic process. 

This is equivalent to defining a Dirichlet process on the space of
stochastic processes, and variations on this idea have been used in a
number of models. This construction was first made explicit in defining the single-p
DDP \cite{MacEachern:1999}. The spatial DP
\cite{Gelfand:Kottas:MacEachern:2004} replaces the stochastic
processes in the single-p DDP with random fields, to create a mixture of surfaces. The
ANOVA DDP \cite{deIorio:Muller:Rosner:MacEachern:2004} creates a
mixture of analysis of variance (ANOVA) to model correlation between
measures associated with categorical covariates. The Linear DDP
\cite{DeIorio:Johnson:Mueller:Rosner:2009} extends the ANOVA DDP to
incorporate a linear term into each of the mixture components, making
the model applicable to continuous data. Since these models can be
interpreted as Dirichlet process mixture models, inference is generally
relatively straightforward.

Since, for most nonparametric random measures found in the literature
(the inhomogeneous beta process \cite{Hjort:1990} being the main
exception), the locations of the random atoms are independent of their
size, this approach can be used with any random measure to create a
dependent random measure. In addition, it can be combined with other
mechanisms that induce dependency in the atom sizes -- for example the
Markov-DDP of \cite{Lin:Grimson:Fisher:2010} and the generalized Polya
urn model of \cite{Caron:Davy:Doucet:2007} both employ this form of
dependency, in addition to mechanisms that allow the size of the atoms
to vary across covariate space.



\section{Stick-breaking constructions}

A large class of nonparametric priors over probability measures of the
form $P:= \sum_{k=1}^\infty\pi_k \delta_{\theta_k}$ can be constructed
using an iterative stick-breaking construction \cite{Sethuraman:1994,Ishwaran:James:2001}, wherein a size-biased
ordering of the atom masses is obtained by repeatedly breaking off
random fractions of a unit length stick, so that
\begin{equation}
\begin{split}
\pi_k &= V_k\prod_{j=1}^{k-1}(1-V_j)\\
V_k&\sim \mbox{Beta}(a_k,b_k) \, .\label{eqn:sbp}
\end{split}
\end{equation}
Many commonly used random probability measures can be represented in
this manner: if we let $a_k = 1, b_k = \gamma$ we recover the
Dirichlet process; if we let $a_k = 1-a, b_k = b+ka$ we recover the
Pitman-Yor, or two-parameter Poisson-Dirichlet process.
A similar procedure has been developed to represent the form of the
beta process most commonly used in the IBP
\cite{Teh:Gor:Gha:2007}. 
A number of
authors have created dependent nonparametric priors by starting from
the stick-breaking construction of Equation~\ref{eqn:sbp}.

\subsection{Varying the beta random variables across covariate space}
 
The multiple-p DDP \cite{MacEachern:2000} replaces the
beta-distributed random variable $V_k$ in Equation~\ref{eqn:sbp} with a stochastic process $V_k(x)$,
whose $x$-marginals are distributed according to $\mbox{Beta}(1,\alpha)$. For
example, $V_k$ might be obtained by point-wise transformation of some
stochastic process whose marginal distribution function is known and
continuous, such as a Gaussian process. The resulting marginal distribution
over random probability measures $G^{(x)}:= \sum_{k=1}^\infty
\pi_k(x)\delta_{ \theta_k}$ are Dirichlet processes by construction. While elegant, inference in the the multiple-p DDP is computationally
daunting, which goes some way to explain why it has not been used in
real-world applications. 


A related model is the kernel stick-breaking process
\citep[KSBP,][]{Dunson:Park:2008}. The KSBP defines a covariate-dependent
mixture $G^{(x)}$ of a countably infinite sequence of probability measures $G^*$ as
\begin{equation*}
G^{(x)} = \sum_{k=1}^\infty V_k K(x, \mu_k)\prod_{j=1}^{k-1}(1- V_j K(x,
\mu_j))G^*_k\; ,
\end{equation*}
where $K\rightarrow[0,1]$ is a bounded kernel function,
$V_k\stackrel{ind}{\sim}\mbox{Beta}(a_k,b_k)$ and $\{\mu_k \in \X\}$ a set of
random covariate locations for the sticks.  If $K=1$ then we recover the class 
of stick-breaking priors; if $K$
varies across $\X$ then the weights in the corresponding marginal
stick-breaking processes vary accordingly.  

While the multiple-p DDP varies the atom weights in such a manner as to
maintain Dirichlet process marginals, the KSBP, in general, does
not. Instead, it modulates the beta-distributed weights using an
arbitrary kernel, resulting in marginally non-beta weights. This model
is much easier to perform inference in than the multiple-p DDP
described above; MCMC schemes have been proposed based on a Polya-urn
representation or using slice sampling. A hierarchical variant of the KSBP has also been proposed, along with
a variational inference scheme \cite{An:Wang:Shterev:Wang:Carin:Dunson:2008}.

The matrix stick-breaking process \cite{Xue:Dunson:Carin:2007} is appropriate for matrix or
array-structured data. Each row $m$ of the matrix is associated with a
sequence $U_{mk}\stackrel{iid}{\sim}\mbox{Beta}(1,\gamma_U),
k=1,2,\dots$, and each column $j$ is associated with a corresponding
sequence $W_{jk}\stackrel{iid}{\sim}\mbox{Beta}(1,\gamma_W)$. At
location $x_{mj}$, corresponding to the $j$th element in the $m$th row
of the matrix, a countable sequence of atom weights is constructed as
\begin{equation}
\pi_{mjk} = U_{mk}W_{jk}\prod_{i=1}^{k-1}(1-U_{mi}W_{ji})\, .
\end{equation}
This model is invariant under permutations of the rows and columns,
and does not depend on any underlying metric. 

\subsection{Changing the order of the beta random variables}

The multiple-p DDP, KSBP, and matrix stick-breaking process all
involve changing the weights of the beta random variables in a
stick-breaking prior. A different approach is followed by Griffin and Steel in constructing the
ordered Dirichlet process, or $\pi$-DDP~\cite{Griffin:Steel:2006}. 

In the $\pi$-DDP, we have a shared set $\{V_k\}_{k=1}^\infty$ of $\mbox{Beta}(1,\alpha)$
random variables and a corresponding set $\{\theta_k\}_{k=1}^\infty$ of random
atom locations. At each covariate value $x \in \X$, we define a permutation
$\sigma^{(x)}$, and let $G^{(x)} = \sum_{k=1}^\infty
\pi_k^{(x)}\delta_{\theta_{\sigma^{(x)}(k)}}$, where $\pi_k^{(x)} =
V_{\sigma^{(x)}(k)}\prod_{j=1}^{k-1}(1-V_{\sigma^{(x)}(j)}$. One method of
defining such a permutation is to associate each of the $(V_k,\theta_k)$ pairs
with a location $\mu_k \in \X$, and taking the permutation implied by the
ordered distances $|\mu_k-x|$.

A related approach is the local Dirichlet process \cite{Chung:Dunson:2011}.
Again, we have shared sets of beta random variables $V_k$, locations in
parameter space $\theta_k$ and locations in covariate space $\mu_k$. In the
local Dirichlet process, for each $x\in\X$ we combine the beta random
variables, and associated parameter locations, for atoms whose covariate
locations are within a neighborhood of $x$, i.e. $|\mu_k-x| < \phi$:
\begin{equation}
G^{(x)} = \sum_{k=1}^{|\mathcal{L}_x|}p_k(x)\delta_{\theta_{\pi_k(x)}}\, ,
\end{equation}
where $\mathcal{L}_x = \{\mu_k: |\mu_k-x| < \phi\}$, $\pi_k(x)$ is the $k$th
ordered index in $\mathcal{L}_x$, and
\begin{equation}
p_k(x) = V_{\pi_k (x)}\prod_{j<k}(1-V_{\pi_j(x)})\, .
\end{equation}

While methods that manipulate the stick-breaking construction are
often elegant, they can be limiting in the form of dependency
available. In the multiple-p DDP and local DP, the size-biased nature
of the stick-breaking process will mean that the general ordering of
the atom sizes (at least for the atoms contributing) will tend to be
similar. In the $\pi$-DDP where the permutation is defined by the
relative distances, atoms are constrained to increase monatonically
with distance to a maximum size and then decrease. In addition,
changes in covariate location will tend to only effect the larger
atoms.


\section{Dependence in conditional distributions}\label{sec:conditional}


According to de Finetti's theorem, any exchangeable sequence is i.i.d.\
given some (latent) conditional distribution, and can be described
using a mixture of such distributions. For example, in the CRP, the
conditional distributions are the class of discrete probability
distributions, and the mixing distribution is a Dirichlet process. In the previous sections, we have discussed ways of inducing dependency in the mixing distribution, and assumed that observations are i.i.d. at each covariate value.

An alternative method of inducing dependency between observations is to assume the mixing distribution is common to all covariate values, but the conditional distributions are correlated. For example, in the Chinese restaurant process, this would mean the mixing measure for all covariate values is shared and distributed according to a single Dirichlet process, and the conditional distributions are correlated but are marginally distributed according to that mixing measure.

The generalized spatial Dirichlet process
\cite{Duan:Guindani:Gelfand:2005, Gelfand:Guindani:Petrone:2007} is an extension of the spatial Dirichlet process \cite{Gelfand:Kottas:MacEachern:2004} that replaces the conditional distribution (in this case, a multinomial) with a collection
of correlated conditional distributions. Recall that the spatial Dirichlet process defines a Dirichlet mixture of Gaussian random fields on some space $\mathcal{X}$. A sample $Y$ from such a process is a realization of the field associated with a single mixture component. 

The generalized spatial Dirichlet process aims to allow a sample to contain
aspects of multiple mixture components. A naive way of achieving this would be
to sample a different field at each location $x\in \X$, but this would not
result is a continuous field over $\X$. Instead, the generalized spatial
Dirichlet process ensures that the value of the field at a given location is
marginally distributed according to the Dirichlet process mixture at that
location, but that as $x\rightarrow x_0$, $p(Y(x) = \theta_i(x), Y(x_0) =
\theta_j(x_0))$ tends to $0$ if $i\neq j$, or to $p(Y(x_0)=\theta_i(x_0))$ otherwise.

This behavior can be achieved as follows. Let $\{Z_i(x), x\in \X,
i=1,2,\dots\}$ be a countable collection of independent Gaussian
random fields on $\X$ with unit variance and mean functions $m_i(x)$
such that $\Phi(m_i(x)) \stackrel{ind.}{\sim}
\mbox{Beta}(1,\gamma)$. Then at each $x\in \X$,  $Y(x) =
\theta_{k(x)}(x), k(x): Z_{k(x)}(x) = \max_i Z_i(x)$.

The idea of using latent surfaces to select mixture components and
hence enforce locally similar clustering structure is explored further
in the latent stick-breaking process
\cite{Roderiguez:Dunson:Gelfand:2010}. In the generalized spatial
Dirichlet process, latent surfaces were combined with a Dirichlet
process distribution over surfaces. In the stick-breaking process,
latent surfaces are used to select locally smooth allocations of
parameters marginally distributed according to an arbitrary
stick-breaking process, and the authors consider multivariate
extensions.  A similar method is used in the dependent Pitman-Yor
process of \cite{Sudderth:Jordan:2008} to segment images. Here, the
authors work in a truncated model and use variational inference
techniques.

A related method is employed in the dependent IBP
\cite{Williamson:Orbanz:Ghahramani:2010}. Recall that, in the beta-Bernoulli
representation of the IBP, we have a random measure $G=\sum_{k=1}^\infty
\pi_k\delta_{\theta_k}$, and each element $z_{nk}$ of a binary matrix
$\mathbf{Z}$ is sampled as $z_{nk}\sim\mbox{Bernoulli}(\pi_k)$. The dependent
IBP couples matrices $\mathbf{Z}(x) = [z_{nk}(x)], x\in \X$ by jointly sampling the elements $z_{nk}(x)$ according to a stochastic process with $\mbox{Bernoulli}(\pi_k)$ marginals. In practice, this is achieved by thresholding a zero-mean Gaussian process with a threshold value of $\Phi^{-1}(\pi_k)$. 



\section{Dependence in predictive distributions} 
\label{sec:predictive}

Recall that when conjugacy exists between the de Finetti mixing measure and the
sampling distribution the predictive distribution $p(X_{n+1}|X_1,\ldots,X_n)$
can in most cases be obtained analytically.  In this section we describe two
approaches to induce dependence using the predictive distribution.  

The first approach
induces dependence in a partially exchangeable\footnote{A set of sequences
$\{X_i=(x_1,x_2,\ldots,x_{n_i})\}$, each with $n_i$ observations, is partially
exchangeable if each sequence is exchangeable, but observations in two
different sequences are not exchangeable.  This is the assumption made by the
two-level hierarchical Dirichlet process and hierarchical beta process and is
appropriate for example modeling text documents where words within a document
are exchangeable but words in different documents are not.} sequence of observations 
that arrive over time in batches by creating Markov chains of CRPs that
incorporate a subset of the observations from the previous time into the current 
CRP. Such approaches maintain CRP-distributed marginals, but are
difficult to extend to covariates other than time.

An alternative approach is to
modify the predictive distribution to explicitly depend on a
covariate (or some function thereof).  This form of construction can be applicable to
sequential or arbitrary covariates.
Unlike many of
the models described elsewhere in this survey, these models are not based on a
single shared random measure.
These models
can be easily adapted to arbitrary covariate spaces, but in general
lack the property of marginal invariance.

\subsection{Markov chains of partitions}

The generalized Poyla urn \citep[GPU,][]{Caron:Davy:Doucet:2007} constructs a 
DDP over time by leveraging the invariance of the combinatorial structure of 
the CRP with respect to subsampling.  Specifically, at time $t=1$ draw a set of
atom assignments for $n_1$ customers, 
$\mathbf{z}_1 = \{z_{1,1},z_{2,1},\ldots,z_{n_1,1}\}$, and associated atoms 
$\mathbf{\theta_1} = \{\theta_{1,1},\ldots,\theta_{K_1,1}\}$
according to a CRP with base measure $G_0$ on $\Theta$, where $K_1$ is the number 
of tables with a customer at time
$1$.  For $t \geq 2$, some subset of the existing customers leave the
restaurant, according to one of two deletion schemes (or a combination):
\begin{enumerate}
\item \textbf{Size-biased deletion}: An entire table is deleted with
  probability proportional to the number of people say at that table.
\item \textbf{Uniform deletion}: Each customer in the restaurant
  decides independently to stay with some probability $q$, otherwise
  they leave.
\end{enumerate}
All remaining atoms that existed 
at time $t-1$, $\theta_{k,t-1}$, are updated by a transition kernel $T(\theta_{k,t} | \theta_{k,t-1})$ 
such that $\int T(\theta_{k,t}|\theta_{k,t-1})G_0(d\theta_{k,t-1}) =
G_0(d\theta_{k,t})$. This is similar to the single-p DDP and related
models described in Section~\ref{sec:var_atoms}.
 
For each of $n_t$ customers at time $t$, sample the seating assignment
for the $i$'th new customer according to a CRP that depends on the customers 
from the previous time step that remained in the restaurant and the $i-1$ new 
customers that entered the restaurant.  Dependence is induced through the
choice of subsampling probability $q$ and the transition kernel
$T(\cdot,\cdot)$. The authors provide sequential Monte Carlo and Gibbs
inference algorithms.

The recurrent Chinese restaurant process \citep[RCRP,][]{Ahmed:Xing:2008} is
similar to, and sometimes a special case of, the GPU.  In the RCRP all
customers leave the restaurant at the end of a time step, however, the atom and
number of customers assigned are remembered for the next time step.  The first
customer to sit at a table from the previous time step is allowed to transition
the associated atom.  The RCRP does not restrict the type of transition to be
invariant to the base measure $G_0$, which means that the marginal measures of
the RCRP may not all be DPs with the same base measure (though they will all be
DPs).  The RCRP can be extended as discussed in \cite{Ahmed:Xing:2008} to model
higher-order correlations by modulating the counts from the previous time by a
decay function \cite{Zhu:Gha:Lafferty:2005}, eg $e^{-h/\lambda}$ where $h$ is a 
lag and $\lambda$ determines the length of influence of the counts at each time.

\subsection{Explicit distance dependency in the predictive
  distribution}
An alternative interpretation of the CRP is to say that customers
choose who to sit next to uniformly, rather than to say customers pick
a table with probability proportional to the table size. An assignment, $\mathbf{c} = \{c_1,\ldots,c_n\}$, of
customers to other customers is equivalent to the usual table-based
interpretation of the CRP by defining two customers $i$ and $j$ to be at the same 
table, ie $z_i = z_j$, if starting at customer $i$ or $j$, there is a sequence of 
customer assignments that starts with $i$ or $j$ and ends with the other, 
eg $(c_i,c_{c_i},\ldots,j)$.  This map is not one-to-one in that the same table
assignment can be generated by two different customer assignments. 

The distance-dependent CRP \citep[ddCRP,][]{Blei:Frazier:2011}
utilizes this representation.  Let $d_{ij}$ 
denote a dissimilarity\footnote{$d_{ij}$ need not satisfy the triangle inequality.} 
measure between customers $i$ and $j$ and let $f(\cdot)$ be a monotonically 
decreasing function of the $d_{ij}$ called a decay function.  Using the customer
assignment interpretation of the CRP, the ddCRP is defined as follows
\begin{equation}
    p(c_i = j | D, \alpha) =
    \begin{cases}
        f(d_{ij}) & \text{if } i \neq j \\
        \alpha & \text{if } i = j
    \end{cases}
    \label{eqn:ddcrp}
\end{equation}
where $\alpha$ is a concentration parameter.  Loosely speaking, a
customer is more likely to choose to sit next to a person he lives near.

Different forms for the decay
function, $f(\cdot)$, allow for the type of dependence to be controlled.  For
example, a ``window'' decay function describes an explicit limit on the maximum
distance between customers that can be assigned to each other.  Soft decay
functions can also be used to allow the influence of customers to diminish over
time.  The ddCRP has been applied to language modeling, a dynamic mixture model
for text and clustering network data \cite{Blei:Frazier:2011}.

This idea of modifying the predictive distribution to
depend on covariates can also be applied to the IBP predictive distribution to create covariate-dependent
latent feature models.  Assume there are $N$ customers and let $s_{ij} =
f(d_{ij})$ denote a 
similarity\footnote{Since $f$ is a monotonically decreasing function of
dissimilarity it can be interpreted as a notion of similarity.} between customers $i$ and $j$ and $w_{ij} = s_{ij} /
\sum_{l=1}^N s_{il}$.

The distance-dependent IBP
\citep[ddIBP,][]{Gershman:Blei:2011} first draws a Poisson number of dishes for
each customer, the dishes drawn for a given customer are said to be ``owned''
by the customer.  For a given dish, $k$, analogously to the ddCRP, each customer chooses
to attach themselves to another customer (possibly themselves) with probability
$w_{ij}$.  After this
process occurs a binary matrix $Z$ is created where entry $z_{ik} = 1$ if
customer $i$ can reach the owner of dish $k$ following the customer
assignments.  Unlike the ddCRP, the direction of the assignments matters:  there must 
exist a sequence of customer assignments starting at $i$ and ending at the dish owner.
If no such path exists then $z_{ik} = 0$.  The ddIBP has been used for
covariate-dependent dimension reduction in a classification task with a latent feature
model.

\section{Dependent Poisson random measures}\label{sec:Poisson}

Since CRMs and Poisson processes are deeply connected (see
Section~\ref{sec:bg}) it is natural to
construct dependent random measures by manipulating the underlying Poisson
process. Recall that a Poisson process on $\Theta \times \mathbb{R}_+$ with rate given by a positive L\'{e}vy measure $\nu(d\theta,d\pi)$, defines a CRM on $\theta$. Therefore, any operation on a Poisson process that yields a new Poisson process will also yield a new CRM. If we ensure that the operation yields a Poisson process with the same rate $\nu$, and allow the operation to depend on some covariate, then we define a dependent CRM that varies with that covariate. From here, we can define a dependent NRM via normalization.

\subsection{Operations on Poisson processes}

In the following, let $\Pi = \{(x_i,\theta_i,\pi_i)\}$ be a Poisson process on
$\X \times \Theta \times \reals_+$ with the product $\sigma$-algebra $\F_\X
\otimes \F_\Theta \otimes \F_{\reals_+}$ \cite{Cinlar:2011} with rate measure
$\nu(dx,d\theta,d\pi)$.  We interpret $\X$ as a space of covariates (with a
metric), $\Theta$ as the parameter space and $\reals_+$ the space of atom
masses. 
Below, we describe three properties of Poisson processes that have been
leveraged to construct dependent CRMs/NRMs: the superposition theorem,
transition theorem, the
mapping theorem, and the restriction theorem. See 
\cite{Kingman:1993,Daley:2003,Daley:2008} for the general statements of these properties.
\paragraph*{\textbf{The superposition theorem}} Let $\Upsilon$ be another
Poisson process on the same space as $\Pi$ with rate measure $\phi(dx, d\theta,
d\pi)$.  Then the superposition, $\Pi \cup \Upsilon$, is again a Poisson
process with updated rate measure $\nu + \phi$.  The superposition follows from
the additive property of Poisson random variables \cite{Kingman:1993}.
Additionally, if $\{\Pi_n\}$ is a countable set of Poisson processes on the
same space with respective rate measures $\{\nu_n\}$, then the superposition,
$\cup_n \Pi_n$, is a Poisson process with rate measure $\sum_n \nu_n$.

\paragraph*{\textbf{The transition theorem}} A transition kernel is a function 
$T: \X \times \Theta \times \F_{\X \times \Theta} \rightarrow [0,1]$, such that
for $(x,\theta) \in \F_{\X \times \Theta}$, $T(x,\theta,\cdot)$ is a
probability measure on $\X \times \Theta$ and for measurable $A$,
$T(\cdot,\cdot,A)$ is measurable.  We denote a sample from $T$ as
$T(x,\theta)$, by which we mean the pair $(x,\theta)$ is moved to the point
$T(x,\theta)$ according to $T$.  The set, $T(\Pi) = \{(T(x,\theta),\pi):
(x,\theta,\pi)\in \Pi\}$ is then a Poisson process with rate $\int_{\X \times
\Theta} T(x,\theta)\nu(dx,d\theta,d\pi)$.

\paragraph*{\textbf{The mapping theorem}} If $\Pi$ is a Poisson process on some 
space $S$ with rate
measure $\nu(\cdot)$, and $f:S\rightarrow T$ is a measureable mapping to some
space $T$, then $f(\Pi)$ is a Poisson process on $T$ with rate measure
$\nu(f^{-1}(\cdot))$. For example, if $S:=\X \times \Theta \times \reals_+$,
and $f_A(\cdot) = \int_{x\in A}\cdot dx$, then $f_A(\Pi)$ is a Poisson process
on $\Theta \times \reals_+$ with rate measure $\int_{x\in A}
\nu(dx,d\theta,d\pi)$. The mapping theorem can be obtained as a
special case of the transition theorem.

\paragraph*{\textbf{Random subsampling}} Let $q : \X \times \Theta \rightarrow
[0,1]$ be a measurable function.  Associate with each atom a Bernoulli random
variable $z_i$ such that $p(z_i = 1) = q(x_i,\theta_i)$.  Then, let $\Pi_b =
\{(x_i,\theta_i,\pi_i) | z_i = b\}$, for $b \in \{0,1\}$ are independent
Poisson processes such that $\Pi_b \sim \PPP(\nu_b)$ where
$\nu_0(dx,d\theta,d\pi) = (1-q(x,\theta))\nu(dx,d\theta,d\pi)$ and
$\nu_1(dx,d\theta,d\pi) = q(x,\theta)\nu(dx,d\theta,d\pi)$ are the respective
rate measures.  The CRM associated with $\Pi_1$ can then be written as
\begin{equation} \label{eqn:thinnedcrm} \sum_{k: z_k=1} \pi_k \delta_{\theta_k}
\end{equation} This is a special case of the marking theorem;  see
\cite{Kingman:1993} and \cite{Daley:2003} for an in-depth
treatment.

\subsection{Superposition and subsampling in a Poisson representation}

A family of bivariate dependent CRMs (and NRMs via normalization) are
constructed in \cite{Lijoi:Nipoti:Pruenster:2011} and \cite{Lijoi:Nipoti:2012} for partially exchangeable sequences of
observations.  The
construction is based on a representation of bivariate Poisson processes due to
Griffiths and Milne \cite{Griffiths:Milne:1978}. If $N_i = M_i
+ M_0$, for $i = 1,2$ where $M_i$ are Poisson processes with rate $\nu$ and
represent idiosyncratic aspects of the $N_i$ and $M_0$ is a baseline Poisson
process with rate measure $\nu$, then by the superposition theorem, the $N_i$ are Poisson processes with rate $2 \nu$. 

This bivariate Poisson process can be used to define a bivariate CRM $(G_1,G_2)$ that takes the form
$G_i = \sum_{k=1}^\infty \pi_{ik}\delta_{\theta_{ik}} + \sum_{k=1}^\infty
\pi_{0k}\delta_{\theta_{0k}}$.  The bivariate CRM, $(G_1,G_2)$ can then be normalized to give a bivariate NRM, $(P_1,P_2)$ which can be used as a prior for a mixture
for partially exchangeable data.  This family of dependent CRMs has been
additionally used as a prior distribution for the hazard rate for partially
exchangeable survival data \cite{Lijoi:Nipoti:2012}.

Lin \textit{et al} \cite{Lin:Grimson:Fisher:2010} use the superposition and
subsampling theorems above to create a discrete-time Markov chain $\Pi_1,\Pi_2\dots$ 
of Poisson processes, and from there, a Markov chain of Dirichlet processes that
we denote the Markov-DDP. Let $\nu(d\theta, d\pi) = \pi^{-1}\exp(-\pi)
d\pi H_0(d\theta)$. At each timestep $i$, the Poisson process $\Pi_{i-1}$ is
subsampled with probability $q$, and is superimposed with an independent
Poisson process with rate $q\nu(d\theta,d\pi)$.
In addition, the atoms from
time $i-1$ are transformed in $\Theta$ using a transition kernel $T$. The
Poisson process at each timestep defines a gamma process, and indeed we can
directly construct the sequence of dependent gamma processes as

\begin{equation} 
    \label{eqn:mccrm} 
    \begin{aligned} 
        G_1 &\sim \text{CRM}(\nu) \\ 
        G_i &= T(S_q(G_{i-1})) + \xi_i \;, i > 1 
    \end{aligned}
\end{equation} 
where $\xi_i \sim \text{CRM}(q\nu)$ and the operation $S_q(G)$
denotes the deletion of each atom of $G$ with probability $q$, corresponding
to subsampling the underlying Poisson process with probability $q$.  The
transition $T$ affects only the locations of the atoms, and not their sizes.
Lin \textit{et al} use the gamma process as the CRM in
Equation~\ref{eqn:mccrm} to generate dependent Dirichlet processes,
but arbitrary CRMs can also be used to generate a wider class of dependent NRMs, as described by Chen
\textit{et al} \cite{Chen:Ding:Buntine:2012}. 

The Markov-DDP is a recharacterization of the size-biased deletion
variant of the GPU DDP described in Section~\ref{sec:predictive},
although the MCMC inference approach used is different. As is clear from Equation~\ref{eqn:mccrm}, we need not instantiate the underlying Poisson process; Lin \textit{et al} employ an MCMC sampler based on the Chinese restaurant process to sample the cluster allocations directly, and Chen \textit{et al} use a slice sampler to perform inference in the underlying CRMs.


There are certain drawbacks to this construction. Firstly, only discrete,
ordered covariates are supported, making it inappropriate for
applications such as spatial modeling.  Second, it is not obvious how
to learn the thinning probability $q$. In the literature $q$ is taken to be a fixed constant and an
ad-hoc method such as cross-validation is needed to find a suitable value.

\subsection{Poisson processes on an augmented space}
An alternative construction of dependent CRMs and NRMs that overcomes these
drawbacks is to use a covariate-dependent kernel function to modulate the
weights of a CRM. This technique was explicitly used in constructing the kernel
beta process, and, as was shown by Foti and Williamson
\cite{Foti:Williamson:2012}, can be used to construct a wider class of
dependent models including the spatial-normalized gamma process
\cite{Rao:Teh:2009}.

As before, let $\Pi = \{(x_i,\theta_i,\pi_i)\}$ be a Poisson process 
on $\X \times \Theta \times \reals_+$ with rate measure $\nu(dx, d\theta, d\pi) = R_0(dx) H_0(d\theta) \nu_0(d\pi)$. Additionally, let $K:\X \times \X \rightarrow [0,1]$ be a bounded kernel function. 
Then, define the set of covariate dependent CRMs $\{G^{(x)} : x \in \X\}$ as
\begin{equation}
    \label{eqn:kcrm}
    G^{(x)} = \sum_{k=1}^\infty K(x,\mu_k) \pi_k \delta_{\theta_k}\, .
\end{equation}
By the mapping theorem for Poisson processes this is a well-defined CRM.

If we take $\nu_0$ to be the L\'{e}vy measure of the homogeneous beta process, we obtain the kernel beta process \cite{Ren:Wang:Dunson:Carin:2011}. Rather than use a single kernel function, the KBP uses a dictionary of exponential kernels with varying widths. 

We can use the kernel beta process to construct a distribution over dependent
binary matrices, by using each $G^{(x)}$ to parameterize a Bernoulli process
(BeP)  $Z^{(x)} \sim \BeP(G^{(x)})$ \cite{Thibaux:Jordan:2007}. As noted in
\cite{Ren:Wang:Dunson:Carin:2011}, using $G^{(x)}$ to parameterize a Bernoulli process has an Indian buffet metaphor where each customer first
decides if they're close enough to a dish in the covariate space ($K(x,\mu_k)$) and if 
so tries the dish with probability proportional to its popularity ($\pi_k$).

The kernel beta process has been used in a covariate dependent factor model
for music segmentation and image denoising and interpolation \cite{Ren:Wang:Dunson:Carin:2011}. 
Inference for the KBP
feature model was performed with a Gibbs sampler on a truncated version of the
measures $G^{(x)}$.

If the L\'{e}vy measure $\nu_0$ in Equation~\ref{eqn:kcrm} is the L\'{e}vy
measure of a gamma process, and we use a box kernel $K(x,\mu) =
\mathbf{I}(||x-\mu||<W)$, then we obtain a dependent gamma process, where each
atom is associated with a location in covariate space and contributes to
measures within a distance $W$ of that location. So, for $x, x' \in \X$, two
gamma processes $G^{(x)}$ and 
$G^{(x')}$ will share more atoms the closer $x$ and $x'$ are in $\X$ and vice
versa. Note that if an atom appears in two measures $G^{(x)}$ and $G^{(x')}$, 
it will have the same mass in each.

If we normalize this gamma process at each covariate value, we obtain the fixed-window form of the spatial normalized gamma process (SNGP) \cite{Rao:Teh:2009}. Placing a prior on the width $W$ allows us to recover the full SNGP model. 

The SNGP can also be obtained using the mapping and subsampling theorems for Poisson processes. Let $\Y$ be an auxiliary space to be made explicit and $\T$ an index set which
we take to be $\reals$ and let $G$ be a gamma process on $\Y \times \Theta$.  
For $t \in \T$ let $Y_t \subset \Y$ be measurable.  
The measure $G^{(t)} = \int_{Y_t} G(dy,d\theta) =
\sum_{k=1}^\infty 1(y_k\in Y_t)\pi_k \delta_{\theta_k}$ is a
gamma process by the mapping
and sub-sampling (using a fixed thinning probability) theorems for
Poisson processes where the rate measure has been updated accordingly.  A set 
of dependent Dirichlet processes is obtained as $D^{(t)} =
G^{(t)}/G^{(t)}(\Theta)$.
As a complete example, suppose that each atom is active for a fixed window of
width $2L$ centered at a time $t$.  In this case $\Y = \reals$ and the sets
$Y_t = [t-L,t+L]$.  In this case, two DPs $D^{(t)}$ and $D^{(t')}$ will share 
atoms as long as
$|s-t| < 2L$.  See \cite{Rao:Teh:2009} for further examples.   The SNGP can be
used with arbitrary covariates, however, covariate spaces of dimension greater
than $2$ become computationally prohibitive because of the geometry of the
$Y_t$.  

A related, but different method for creating dependent NRMs using time as the
underlying covariate was presented in \cite{Griffin:2011}.  Let $\Pi =
\{(\mu_i,\theta,\pi_i)\}$ be a Poisson process on $\reals \times \Theta \times
\reals_+$ with a carefully constructed rate measure $\nu(dt,d\theta,d\pi)$.
Then, a family of time-varying NRMs $\{G_t, t\in\reals\}$ can be constructed where
\begin{equation}
    \label{eqn:ounrm}
    G_t = \sum_{k=1}^\infty \frac{I(\mu_k \leq
    t)\exp(-\lambda(t-\mu_k))\pi_k}{\sum_{l=1}^\infty I(\mu_l \leq
    t)\exp(-\lambda(t-\mu_l))\pi_l}\delta_{\theta_k}
\end{equation}
This family of time-varying NRMs are denoted Ornstein-Uhlenbeck NRMS (OUNRM), since the kernel used to
define $G_t$ is an Ornstein-Uhlenbeck kernel.  Though similar to the KNRMs
presented above, the L\'{e}vy measure used to construct an OUNRM utilizes
machinery for stochastic differential equations in order to ensure that the
marginal measures, $G_t$, are of the same form as the original NRM on the
larger space \cite{BNS:2001}.  In particular the OUNRM construction makes use of the
Ferguson-Klass representation \cite{FK:1972} of a stochastic integral with respect to a
L\`{e}vy process which dictates the form of $\nu$ above.  
Inference can be performed with an MCMC
algorithm that utilizes Metropolis-Hastings steps as well as slice sampling
\cite{Damien:1999}.  In addition, a particle filtering algorithm
\cite{Doucet:2001} has been derived to perform online inference.  

Another method that makes use of the Ferguson-Klass representation is
the Poisson line process-based dependent CRM
\cite{Williamson:2012}. A Poisson line process is a Poisson process on
the space of infinite straight lines in $\reals^2$ \cite{Miles:1964}, and
so a sample from a Poisson line process is a collection of straight
lines in the plane, such that the number of lines passing through a
convex set is Poisson-distributed with mean proportional to the
measure of that set.

A useful fact of Poisson line processes is that the intersections of a
homogeneous Poisson line process with an arbitrary line in $\reals^2$
describes a homogeneous Poisson point process on $\reals$. Clearly,
the intersections of a Poisson line process with two lines
$\ell,\ell'\in\reals^2$ describe a dependent Poisson point process - marginally,
each describes a Poisson point process, but the two processes are
correlated. The closer the two lines, the greater the correlation
between the Poisson processes. 

In
\cite{Williamson:2012}, covariate values are mapped to lines in
$\reals^2$, and a Poisson line process on $\reals^2$ induces a collection of
dependent Poisson processes corresponding to those values. Judicious
choice of the rate measure of the Poisson line process ensures that
the marginal Poisson processes on $\reals$ at each covariate value have
rate $1/2$. Taking the absolute value of these Poisson processes with
respect to some origin gives a unit-rate Poisson process, which can be
transformed to give an arbitrary CRM (including inhomogeneous
variants) via the Ferguson-Klass
representation as described in Section~\ref{sec:CRM}.

\section{Superpositions and convex combinations of random measures}

An alternative way of looking at the SNGP is as a
convex combination of Dirichlet processes. In the Poisson process
representation of the SNGP, described in Section~\ref{sec:Poisson}, a Poisson process is
constructed on an augmented space. At each covariate location $x\in\X$, we
obtain a covariate-dependent Poisson process by restricting this large Poisson
process to a subset $A_x$ of the augmented space.

For a finite number of covariate locations, the corresponding subsets define a
finite algebra. Let $\mathcal{R}$ be the smallest collection of disjoint
subsets in this algebra such that each $A_x$ is a union of subsets in
$\mathcal{R}$. Each subset $r\in\mathcal{R}$ is associated with an
\emph{independent} Poisson process, and hence an independent gamma process
$G^{(r)}$. These gamma processes can be represented as unnormalized Dirichlet
processes, with gamma-distributed weights. Each $A_x$ is therefore associated
with a gamma process $G^{(x)}$ obtained by the \emph{superposition} of the
gamma processes $G^{(r)}:r\in A_x \cap \mathcal{R}$. If we normalize the gamma process $G^{(r)}$ we obtain a \emph{mixture} of Dirichlet processes, with Dirichlet-distributed weights corresponding to the normalized masses of the $G^{(r)}$. Similarly, the bivariate random measures of \cite{Lijoi:Nipoti:Pruenster:2011, Lijoi:Nipoti:2012} and the partially exchangeable model of \cite{Muller:Quintana:Rosner:2004} can be represented as superpositions or convex combinations of random measures.

A number of other models have been obtained via convex combinations of
random measures associated with locations in covariate space, although
these models do not necessarily fulfill the desiderata of having
marginals distributed according to a standard nonparametric process.

The dynamic Dirichlet process \cite{Dunson:2006} is an autoregressive
extension to the Dirichlet process. The measure at time-step $i$ is a
convex combination of the measure at time-step $i-1$ and a DP-distributed
innovation measure:
\begin{equation}
\begin{split}
G_1 &\sim \DP(\gamma_0, G_0)\\
H_i &\sim \DP(\gamma_i,H_{0i})\\
\tilde{w}_i &\sim \mbox{Beta}(a_{w(i)}, b_{w(i)})\\
G_i &= (1-\tilde{w}_{i-1})G_{i-1} + \tilde{w}_{i-1} H_{i-1}, i>1.
\end{split}
\end{equation}
The measure at each time step is this a convex combination of
Dirichlet processes. Note that, in general, this will not be marginally
distributed according to a Dirichlet process, and that the marginal
distribution will depend on time.

The dynamic hierarchical Dirichlet process
\cite{Ren:Dunson:Carin:2008} extends this model to grouped data. Here,
we introduce the additional structure $H_{0i}:=G_0\sim \DP(\gamma,
H)$.  We see that each measure $G_t$ is a convex combination of $G_1,
H_1,\dots, H_{t-1}$, and that these basis measures are samples from a
single HDP. Again, unlike the SNGP, this model does not have Dirichlet process-distributed marginals, and can only be applied along a single discrete, ordered covariate.

A related model that is applicable to more general covariate spaces is the
Bayesian density regression model of \cite{Dunson:Pillai:Park:2007}. Here, the
random probability measure $G^{(x_i)}$ at each covariate value $x_i$ is given by a convex combination of the measures at neighbouring covariate values plus an innovation random measure:

\begin{equation*}
G^{(x_i)} = a_{ii} G^{*(x_i)} + \sum_{j \sim i}a_{ij}G^{(x_j)}
\end{equation*}
where $(j\sim i)$ indicates the set of locations within a defined neighborhood
of $x_i$. A similar approach has been applied to the beta process to create a dependent hierarchical beta process (dHBP) \cite{Zhou:Yang:Sapiro:Dunson:Carin:2011}.

\section{Applications}\label{sec:applications}

Dependent nonparametric processes were originally proposed to flexibly model the
error distribution in a regression model, $y_i = f(x_i) + \epsilon_i$
\cite{MacEachern:1999}.  Whereas traditionally the distribution of $\epsilon_i$ 
is Gaussian or some other parametric form, using a dependent stochastic process
the $\epsilon_i$ can be distributed according to an arbitrary distribution
$F_{x_i}$ that may depend on the observed covariate, $x_i$. 

Early research into dependent nonparametric processes focused on the
use of dependent Dirichlet processes (or Dirichlet-based processes)
for use in regression and density estimation. The survey by
Dunson \cite{Dunson:2010} provides a thorough overview of covariate-dependent 
density estimation. Most of this early work originates from the
statistics literature.

The breadth of applications expanded rapidly once the machine learning
community began to realize the potential of dependent nonparametric
processes. In this section, we describe some recent machine learning applications of covariate-dependent 
nonparametric Bayesian models.

\subsection{Image processing}
A number of dependent stochastic processes have been applied to image
processing applications, including denoising, inpainting
(interpolation), and image segmentation.

In the denoising problem a noisy version of the image is provided and the goal
is to uncover a smoothed version with the noise removed.  In image inpainting 
only a fraction of the pixels are
actually observed and the goal is to impute the values of missing pixels under
the learned model.  In both cases a common pre-processing step is to break the
image up into small patches of dimension $m \times m$ ($m = 8$ is frequently
used) that are often overlapping.  Each patch is then treated as a vector $y_i
\in \reals^{m^2}$.

The most common Bayesian nonparametric model for these
problems is a sparse factor model where $y_i = Dw_i + \epsilon_i$, where the
columns of $D \in \reals^{m^2 \times \infty}$ are the dictionary elements (or
factors), $w
\in \reals^\infty$ are the factor weights and $\epsilon_i \sim
N(0,\sigma_\epsilon)$.  
are decomposed as $w_i = z_i \odot s_i$ where $z_i$ is a binary vector and the
entries of $s_i \sim N(0,\sigma_s)$.  %
In a stationary model, the entries of $z_i$ are typically distributed
according to an Indian buffet process, or equivalently their latent
probabilities are described using a beta process.
\cite{Zhou:2012:BPFA}. However, this makes the poor assumption that
image patches are exchangeable.

A better approach is to use a dependent nonparametric prior to
generate the feature probabilities, since in addition to the
expectation of global structure, each patch overlaps with its
neighbors, thus sharing local structure. This form of structure can be
achieved using the dHBP and KBP; using such models has achieved
superior denoising and more compact dictionaries
\cite{Zhou:Yang:Sapiro:Dunson:Carin:2011,Ren:Wang:Dunson:Carin:2011}.

Another important and challenging problem in image processing is
segmenting images into coherent regions, such as ``sky'', ``house'', or
``dog''. This is essentially a structured clustering task - each
superpixel belongs to a single cluster, but proximal superpixels are
likely to have similar assignments.

This is exactly the sort of structure achieved by the latent
stick-breaking processes described in
Section~\ref{sec:conditional}. The dependent Pitman-Yor process has
been used to achieve state-of-the art segmentation of natural scenes \cite{Sudderth:Jordan:2008}.


\subsection{Music segmentation}
A common task in analyzing music is determining segments of a song that are
highly correlated in time.  One way of modeling the evolution of music
is to use a hidden Markov model (HMM) to model the Mel frequency cepstral coefficients
(MFCCs) of a piece of music across time. However, such a model does not
allow for evolution of the transition distribution accross
time. Better results can be obtained by using a dymanic HDP as the
basis of the HMM, as described by \cite{Ren:Wang:Dunson:Carin:2011}. 

Another approach is to use a sparse factor model \cite{Zhou:2012:BPFA}
in a manner analogous to the image segmentation problem
above. Obviously, a stationary model will miss local correlations,
suggesting the use of a dependent model. The KBP was able to achieve
better segmentation than both the
stationary BPFA model
\cite{Zhou:2012:BPFA} and the dymanic HDP-HMM \cite{Ren:Dunson:Lindroth:Carin:2010}, 
both of which learned a blockier correlation matrix.

\subsection{Topic modeling}
Topic modeling is a class of techniques for modeling documents as
exchangeable collections of words drawn from a document-specific
distribution over a global set of ``topics'', or distributions over
words, for the purpose of decomposing a collection of text
documents into the underlying topics. The canonical topic model is
latent Dirichlet allocation 
\citep[LDA,][]{Blei:2003}, a stationary
model employing a finite number of topics. The HDP has allowed the
construction of nonparametric topic models with an unbounded number of
topics \cite{Teh:Jordan:Beal:Blei:2006}.

Many corpora evolve over time - for example news archives or
journal volumes. Standard topic models such as LDA and topic models based on the HDP
assume that documents are exchangeable, but in fact we may expect to
see changes in the topic distribution over time, as topics wax and
wane in probability and the language used to describe a topic changes
over time.

A number of dependent Dirichlet processes have been used to create
time-dependent topic models.  For example, the SNGP \cite{Rao:Teh:2009} has been used to construct a 
dynamic HDP 
topic model.  The marginal DP at time $t$ is used as the base measure
in an HDP that in turn models the
topic distribution used by the documents at time $t$.  Since atoms are only
active for a finite window, each topic will appear for only a finite
amount of time before disappearing. The generalized Polya urn model of
\cite{Caron:Davy:Doucet:2007} has been used in a similar setting; by
using the uniform deletion formulation of the model, topics are
allowed to increase and decrease in popularity multiple times.

In the Markov-DDP of \cite{Lin:Grimson:Fisher:2010}, in addition to
varying the topic probabilities, the topics  themselves, which are just
(finite) probability distributions over a fixed vocabulary, are allowed to
evolve over time. The recurrent Chinese restaurant process is used in
a similar manner in the Timeline model of \cite{Ahmed:Xing:2010}


\subsection{Financial applications}
Dependent nonparametric processes can also be applied to problems in finance.
The volatility of the price of a financial instrument is a measure of the
instrument's price variation.  High volatility implies large changes in price
and vice versa.  Modeling volatility is a challenging problem, see
\cite{Tsay:2010} for an overview.  A common model choice is a stochastic 
volatility model, a simple version of which is to model
the log-returns, $r_t$ definied as $\log P_t - \log P_{t-1}$ for $P_t$ the price at
time $t$, as $r_t \sim N(0,\sigma_t)$.  The choice of the distribution for
$\sigma_t$ determines the type of stochastic volatility model.  The OUNRM has
been used in this setting \cite{Griffin:2011} where we model $\sigma_t \sim G_t$ 
and where $G_t \sim \text{OUNRM}$ is drawn from an OUNRM process with an 
inverse-gamma distribution base measure to ensure a positive variance.
The results reported in \cite{Griffin:2011} indicate that the time-dependent 
nonparametric modeling of the volatility fit actual volatility well and by explicitly
modeling the time-dependence the model is able to estimate the length of the
effects of shocks and other interesting phenomena.

\section{Conclusion}

In this paper, we have attempted to provide a snapshot of the current
range of dependent nonparametric processes employed by the machine
learning and statistics communities. Our goal was to curate the wide
range of papers on this topic into a few classes to highlight the main
methods being used to induce dependency between random measures. We
hope that, by highlighting the links between seemingly disparate
models, the practicioner will find it easier to navigate the available
models or develop new models that are well matched to the application
at hand. We also hope that, by realizing and leveraging similarities between related
models, the practitioner will find it easier to identify an efficient
and appropriate inference technique for the model at hand.

While the history of dependent stochastic processes is relatively
short, modern models have moved well beyond the single-p DDP
introduced in \cite{MacEachern:2000}. Early research in the field was
driven by the statistics community, and was mostly focused on
classical statistical problems such as density
estimation and regression and adhered to the theoretical desiderata of
MacEachern reproduced in Section~\ref{sec:DDPintro}.

As the machine learning community has woken up to the potential of
dependent nonparametric processes, the range of applications
considered has ballooned. In particular, the use of dependent
nonparametric models has been active in areas such as text and image
analysis. This expansion has also seen an increase in more scaleable
inference algorithms based on variational or sequential approaches.

However, the models introduced by the machine learning community have
often exhibited less theoretical rigor. Many of the processes developed in machine learning do not meet all of the
desiderata of MacEeachern.  For instance many of these processes do
not have known marginals and some are not even stationary.

There is certainly an argument that the definition put forth by
MacEachern is overly restrictive. While models such as
the KSBP do not have marginals directly corresponding to known
nonparametric priors, they are still useful and well defined
priors. And while models such as the ddCRP and the dHBP rely on having
a known set of covariate locations and may lack well-defined
out-of-sample prediction methods, there are many applications - such
as image denoising - where out-of-sample prediction is not an
important task. 

The question arises of whether the framework of MacEachern is still applicable for
modern dependent processes.  Certainly large support of the prior, efficient
inference and posterior consistency are still relevant.  However, playing (relatively) fast and loose with the
theoretical properties of models has often led to more manageable
inference algorithms. We feel that the MacEachern framework is still a
useful starting point, but believe that more emphasis should be placed on
tractable inference, and on ensuring the dependence between $\{G^{(x)} :
x \in \X\}$ is appropriate for the data, rather than using an overly
restrictive form of dependence for the sake of theory.

Conversely, we hope to see a more rigorous analysis of the
theoretical properties of dependent nonparametric processes,
particularly from the machine learning community. We appeal to the statistics community to develop underlying theory 
for the processes from machine learning to
provide confidence when using them. 
The successful development
of dependent nonparametric processes to their fullest potential depends on the 
complementary interests and expertise of the statistics and machine learning 
communities.

\bibliographystyle{unsrt}
\bibliography{paper}

\begin{thebibliography}{10}

\bibitem{MacEachern:2000}
S.~N. MacEachern.
\newblock Dependent {D}irichlet processes.
\newblock Technical report, Department of Statistics, Ohio State University,
  2000.

\bibitem{Ferguson:1973}
T.~S. Ferguson.
\newblock A {B}ayesian analysis of some nonparametric problems.
\newblock {\em Annals of Statistics}, 1(2):209--230, 1973.

\bibitem{Pitman:2002}
J.~Pitman.
\newblock Combinatorial stochastic processes.
\newblock In J.~Picard, editor, {\em \'{E}cole d'\'{E}t\'{e} de
  Probabilit\'{e}s de Saint-Flour XXXII}, number 1875 in Lecture notes in
  mathematics. Springer, 2002.

\bibitem{Perman:Pitman:Yor:1992}
M.~Perman, J.~Pitman, and M.~Yor.
\newblock Size-biased sampling of {P}oisson point processes and excursions.
\newblock {\em Probability Theory and Related Fields}, 92(1):21--39, 1992.

\bibitem{Griffiths:Ghahramani:2005}
T.~L. Griffiths and Z.~Ghahramani.
\newblock Infinite latent feature models and the {I}ndian buffet process.
\newblock In {\em Advances in Neural Information Processing Systems}, 2005.

\bibitem{Tsay:2010}
R.~S. Tsay.
\newblock {\em Analysis of Financial Time Series}.
\newblock Wiley Series in Probability and Statistics. John Wiley \& Sons, 2010.

\bibitem{Cressie:2011}
N.~A.~C. Cressie and C.~K. Wikle.
\newblock {\em Statistics for Spatio-Temporal Data}.
\newblock Wiley Series in Probability and Statistics. Wiley, 2011.

\bibitem{Cifarelli:Regazzini:1978}
D.~M. Ciffarelli and E.~Regazzini.
\newblock Nonparametric statistical problems under partial exchangeability. the
  use of associative means.
\newblock {\em Annali del'Instituto di Matematica Finianziaria dell'Universita
  di Torino}, 3(12):1--36, 1978.

\bibitem{Lin:Grimson:Fisher:2010}
D.~Lin, E.~Grimson, and J.~Fisher.
\newblock Construction of dependent {D}irichlet processes based on {P}oisson
  processes.
\newblock In {\em Advances in Neural Information Processing Systems}, 2010.

\bibitem{Caron:Davy:Doucet:2007}
F.~Caron, M.~Davy, and A.~Doucet.
\newblock Generalized {P}olya urn for time-varying {D}irichlet process
  mixtures.
\newblock In {\em Uncertainty in Artificial Intelligence}, 2007.

\bibitem{Griffin:2011}
J.~E. Griffin.
\newblock The {O}rnstein-{U}hlenbeck {D}irichlet process and other time-varying
  processes for {B}ayesian nonparametric inference.
\newblock {\em Journal of Statistical Planning and Inference}, 141(11):3648 --
  3664, 2011.

\bibitem{Williamson:2012}
S.~A. Williamson.
\newblock {\em Bayesian nonparametric models for dependent data}.
\newblock PhD thesis, University of Cambridge, 2011.

\bibitem{Williamson:Orbanz:Ghahramani:2010}
S.~Williamson, P.~Orbanz, and Z.~Ghahramani.
\newblock Dependent {I}ndian buffet processes.
\newblock In {\em International Conference on Artificial Intelligence and
  Statistics}, 2010.

\bibitem{Zhou:Yang:Sapiro:Dunson:Carin:2011}
M.~Zhou, H.~Yang, G.~Sapiro, D.~B. Dunson, and L.~Carin.
\newblock Dependent hierarchical beta process for image interpolation and
  denoising.
\newblock In {\em International Conference on Artificial Intelligence and
  Statistics}, 2011.

\bibitem{Ren:Wang:Dunson:Carin:2011}
L.~Ren, Y.~Wang, D.~B. Dunson, and L.~Carin.
\newblock The kernel beta process.
\newblock In {\em Advances in Neural Information Processing Systems}, 2011.

\bibitem{Dunson:2010}
D.~B. Dunson.
\newblock Nonparametric {B}ayes applications to biostatistics.
\newblock In {\em Bayesian Nonparametrics: Principles and Practice}, 2010.

\bibitem{Saeedi:2011}
A.~Saeedi and A.~Bouchard-C\^ot\'e.
\newblock Priors over recurrent continuous time processes.
\newblock In {\em Advances in Neural Information Processing Systems}, 2011.

\bibitem{Kingman:1967}
J.~F.~C. Kingman.
\newblock Completely random measures.
\newblock {\em Pacific Journal of Mathematics}, 21(1):59--78, 1967.

\bibitem{FK:1972}
T.~S. Ferguson and M.~J. Klass.
\newblock A representation of independent increment processes without
  {G}aussian components.
\newblock {\em Annals of Mathematical Statistics}, 43(5):1634--1643, 1972.

\bibitem{Orbanz:Williamson:2011}
P.~Orbanz and S.~Williamson.
\newblock Unit-rate poisson representations of completely random measures.
\newblock Technical report, 2012.

\bibitem{Lijoi:Pruenster:2010}
A.~Lijoi and I.~Pr\"{u}nster.
\newblock Models beyond the {D}irichlet process.
\newblock In N.~L. Hjort, C.~Holmes, P.~M\"{u}ller, and S.~G. Walker, editors,
  {\em Bayesian Nonparametrics}, pages 80--136. Cambridge University Press,
  2010.

\bibitem{Lijoi:Pruenster:Regazzini:2003}
A.~Lijoi, I.~Pr\"{u}nster, and E.~Regazzini.
\newblock Distributional results for means of normalized random measures with
  independent increments.
\newblock {\em Annals of Statistics}, 31(2):560--585, 2003.

\bibitem{Teh:Jor:2010a}
Y.~W. Teh and M.~I. Jordan.
\newblock Hierarchical {B}ayesian nonparametric models with applications.
\newblock In N.~Hjort, C.~Holmes, P.~M{\"u}ller, and S.~Walker, editors, {\em
  Bayesian Nonparametrics: Principles and Practice}. Cambridge University
  Press, 2010.

\bibitem{Aldous:1983}
D.~J. Aldous.
\newblock Exchangeability and related topics.
\newblock In {\em \'{E}cole d'\'{E}t\'{e} de Probabilit\'{e}s de Saint-Flour
  XIII}, pages 1--198. 1983.

\bibitem{Thibaux:Jordan:2007}
R.~Thibaux and M.~I. Jordan.
\newblock Hierarchical beta processes and the {I}ndian buffet process.
\newblock In {\em International Conference on Artificial Intelligence and
  Statistics}, 2007.

\bibitem{Knowles:Ghahramani:2007}
D.~A. Knowles and Z.~Ghahramani.
\newblock Infinite sparse factor analysis and infinite independent components
  analysis.
\newblock In {\em ICA}, 2007.

\bibitem{Zhou:Hannah:Dunson:Carin:2012}
M.~Zhou, L.~A. Hannah, D.~B. Dunson, and L.~Carin.
\newblock Beta-negative binomial process and {P}oisson factor analysis.
\newblock In {\em International Conference on Artificial Intelligence and
  Statistics}, 2012.

\bibitem{Titsias:2007}
M.~K. Titsias.
\newblock The infinite gamma-{P}oisson feature model.
\newblock In {\em Advances in Neural Information Processing Systems}, 2007.

\bibitem{Teh:Jordan:Beal:Blei:2006}
Y.~W. Teh, M.~I. Jordan, M.~J. Beal, and D.~M. Blei.
\newblock Hierarchical {D}irichlet processes.
\newblock {\em Journal of the American Statistical Association},
  101(476):1566--1581, 2006.

\bibitem{Muller:Quintana:Rosner:2004}
P.~M\"{u}ller, F.~Quintana, and G.~Rosner.
\newblock A method for combining inference across related nonparametric
  {B}ayesian models.
\newblock {\em Journal of the Royal Statistical Society: Series B},
  66(3):735--749, 2004.

\bibitem{Fox:2008}
E.~B. Fox, E.~B. Sudderth, M.~I. Jordan, and A.~S. Willsky.
\newblock An {HDP-HMM} for systems with state persistence.
\newblock In {\em Proc. International Conference on Machine Learning}, July
  2008.

\bibitem{Xing:Soh:Jor:Teh:2006}
E.~P. Xing, K.-A. Sohn, M.~I. Jordan, and Y.~W. Teh.
\newblock {B}ayesian multi-population haplotype inference via a hierarchical
  {D}irichlet process mixture.
\newblock In {\em Proceedings of the International Conference on Machine
  Learning}, volume~23, 2006.

\bibitem{Teh:Kur:Wel:2008}
Y.~W. Teh, K.~Kurihara, and M.~Welling.
\newblock Collapsed variational inference for {HDP}.
\newblock In {\em Advances in Neural Information Processing Systems},
  volume~20, 2008.

\bibitem{Wang:Paisley:Blei:2011}
C.~Wang, J.~W. Paisley, and D.~M. Blei.
\newblock Online variational inference for the hierarchical dirichlet process.
\newblock {\em Journal of Machine Learning Research - Proceedings Track},
  15:752--760, 2011.

\bibitem{Bryant:Sudderth:2012}
M.~Bryant and E.~Sudderth.
\newblock Truly nonparametric online variational inference for hieriarchical
  dirichlet processes.
\newblock In {\em Advances in Neural Information Processing Systems}, 2012.

\bibitem{Doucet:2001}
A.~Doucet, N.~de~Freitas, N.~Gordon, and A.~Smith.
\newblock {\em Sequential Monte Carlo Methods in Practice}.
\newblock Statistics for Engineering and Information Science. Springer, 2001.

\bibitem{Ahmed:2011}
A.~Ahmed, Q.~Ho, C.~H. Teo, J.~Eisenstein, A.~J. Smola, and E.~P. Xing.
\newblock Online inference for the infinite topic-cluster model: Storylines
  from streaming text.
\newblock {\em Journal of Machine Learning Research - Proceedings Track},
  15:101--109, 2011.

\bibitem{MacEachern:1999}
S.~N. MacEachern.
\newblock Dependent nonparametric processes.
\newblock In {\em Proceedings of the Section on Bayesian Statistical Science},
  1999.

\bibitem{Gelfand:Kottas:MacEachern:2004}
A.~E. Gelfand, A.~Kottas, and S.~N. MacEachern.
\newblock Bayesian nonparametric spatial modelling with {D}irichlet process
  mixing.
\newblock {\em Journal of the American Statistical Association},
  100:1021--1035, 2005.

\bibitem{deIorio:Muller:Rosner:MacEachern:2004}
M.~{De Iorio}, P.~Muller, G.~L. Rosner, and S.~N. MacEachern.
\newblock An {ANOVA} model for dependent random measures.
\newblock {\em Journal of the American Statistical Association}, 99:205--215,
  2004.

\bibitem{DeIorio:Johnson:Mueller:Rosner:2009}
M.~De~Iorio, W.~O. Johnson, P.~M\"{u}ller, and G.~L. Rosner.
\newblock Bayesian nonparametric nonproportional hazards survival modeling.
\newblock {\em Biometrics}, 65(3):762--771, 2009.

\bibitem{Hjort:1990}
N.~L. Hjort.
\newblock Nonparametric {B}ayes estimators based on beta processes in models
  for life history data.
\newblock {\em Annals of Statistics}, 18(3):1259--1294, 1990.

\bibitem{Sethuraman:1994}
J.~Sethuraman.
\newblock A constructive definition of {D}irichlet priors.
\newblock {\em Statistica Sinica}, 4(2):639--650, 1994.

\bibitem{Ishwaran:James:2001}
H.~Ishwaran and L.~F. James.
\newblock Gibbs sampling methods for stick-breaking priors.
\newblock {\em Journal of the American Statistical Association},
  96(453):161--173, 2001.

\bibitem{Teh:Gor:Gha:2007}
Y.~W. Teh, D.~{G\"or\"ur}, and Z.~Ghahramani.
\newblock Stick-breaking construction for the {I}ndian buffet process.
\newblock In {\em International Conference on Artificial Intelligence and
  Statistics}, 2007.

\bibitem{Dunson:Park:2008}
D.~B. Dunson and J.-H. Park.
\newblock Kernel stick-breaking processes.
\newblock {\em Biometrika}, 95(2):307--323, 2008.

\bibitem{An:Wang:Shterev:Wang:Carin:Dunson:2008}
Q.~An, C.~Wang, I.~Shterev, E.~Wang, L.~Carin, and D.~B. Dunson.
\newblock Hierarchical kernel stick-breaking process for multi-task image
  analysis.
\newblock In {\em International Conference on Machine Learning}, 2008.

\bibitem{Xue:Dunson:Carin:2007}
Y.~Xue, D.~B. Dunson, and L.~Carin.
\newblock The matrix stick-breaking process for flexible multi-task learning.
\newblock In {\em International Conference on Machine Learning}, 2007.

\bibitem{Griffin:Steel:2006}
J.~E. Griffin and M.~F.~J. Steel.
\newblock Order-based dependent {D}irichlet processes.
\newblock {\em Journal of the American Statistical Association}, 101:179--194,
  2006.

\bibitem{Chung:Dunson:2011}
Y.~Chung and D.~B. Dunson.
\newblock The local {D}irichlet process.
\newblock {\em Annals of the Institute for Statistical Mathematics},
  (63):59--80, 2011.

\bibitem{Duan:Guindani:Gelfand:2005}
A.~Duan, M.~Guindani, and A.~E. Gelfand.
\newblock Generalized spatial {D}irichlet process models.
\newblock {\em Biometrika}, 94(4):809--825, 2007.

\bibitem{Gelfand:Guindani:Petrone:2007}
A.~E. Gelfand, M.~Guindani, and S.~Petrone.
\newblock Bayesian nonparametric models for spatial data using dirichlet
  processes.
\newblock {\em Bayesian Statistics}, 8:1--26, 2007.

\bibitem{Roderiguez:Dunson:Gelfand:2010}
A.~Roder\'{i}guez, D.~B. Dunson, and A.~E. Gelfand.
\newblock Latent stick-breaking processes.
\newblock {\em Journal of the American Statistical Association},
  105(490):647--659, 2010.

\bibitem{Sudderth:Jordan:2008}
E.~B. Sudderth and M.~I. Jordan.
\newblock Shared segmentation of natural scenes using dependent {P}itman-{Y}or
  processes.
\newblock In {\em Advances in Neural Information Processing Systems}, 2008.

\bibitem{Ahmed:Xing:2008}
A.~Ahmed and E.~P. Xing.
\newblock Dynamic non-parametric mixture models and the recurrent chinese
  restaurant process: with applications to evolutionary clustering.
\newblock In {\em SIAM International Conference on Data Mining}, 2008.

\bibitem{Zhu:Gha:Lafferty:2005}
X.~Zhu, Z.~Ghahramani, and J.~Lafferty.
\newblock {Time-sensitive Dirichlet process mixture models}.
\newblock Technical report, Carnegie Mellon University, 2005.

\bibitem{Blei:Frazier:2011}
D.~M. Blei and P.~I. Frazier.
\newblock Distance dependent {C}hinese restaurant processes.
\newblock {\em Journal of Machine Learning Research}, 12:2461--2488, 2011.

\bibitem{Gershman:Blei:2011}
S.~J. Gershman, P.~I. Frazier, and D.~M. Blei.
\newblock Distance dependent infinite latent feature models.
\newblock arXiv:1110.5454, 2011.

\bibitem{Cinlar:2011}
E.~Cinlar.
\newblock {\em Probability and Stochastics}.
\newblock Graduate Texts in Mathematics. Springer, 2011.

\bibitem{Kingman:1993}
J.~F.~C. Kingman.
\newblock {\em Poisson Processes}.
\newblock Oxford University Press, 1993.

\bibitem{Daley:2003}
D.~J. Daley and D.~Vere-Jones.
\newblock {\em {An Introduction to the Theory of Point Processes, Volume 1 (2nd
  ed.)}}.
\newblock Springer, New York, 2003.

\bibitem{Daley:2008}
D.~J. Daley and D.~Vere-Jones.
\newblock {\em An introduction to the theory of point processes. {V}ol. {II}}.
\newblock Probability and its Applications (New York). Springer, second
  edition, 2008.

\bibitem{Lijoi:Nipoti:Pruenster:2011}
A.~Lijoi, B.~Nipoti, and I.~Pr\"{u}nster.
\newblock Bayesian inference with dependent normalized completely random
  measures.
\newblock Technical report, Collegio Carlo Alberto, 2011.

\bibitem{Lijoi:Nipoti:2012}
A.~Lijoi and B.~Nipoti.
\newblock A class of dependent random hazard rates.
\newblock Technical report, Collegio Carlo Alberto, 2012.

\bibitem{Griffiths:Milne:1978}
R.~C. Griffiths and R.~K. Milne.
\newblock A class of bivariate {P}oisson processes.
\newblock {\em Journal of Multivariate Analysis}, 8:380--395, 1978.

\bibitem{Chen:Ding:Buntine:2012}
C.~Chen, N.~Ding, and W.~L. Buntine.
\newblock Dependent hierarchical normalized random measures for dynamic topic
  modeling.
\newblock {\em International Conference on Machine Learning}, 2012.

\bibitem{Foti:Williamson:2012}
N.~Foti and S.~Williamson.
\newblock Slice sampling dependent normalized kernel-weighted completely random
  measure mixture models.
\newblock In {\em Advances in Neural Information Processing Systems}, 2012.

\bibitem{Rao:Teh:2009}
V.~Rao and Y.~W. Teh.
\newblock Spatial normalized gamma processes.
\newblock In {\em Advances in Neural Information Processing Systems}, 2009.

\bibitem{BNS:2001}
O.~E. Barndorff-Nielsen and N.~Shephard.
\newblock Non-{G}aussian {O}rnstein-{U}hlenbeck-based models and some of their
  uses in financial economics.
\newblock {\em Journal of the Royal Statistical Society Series B},
  63(2):167--241, 2001.

\bibitem{Damien:1999}
P.~Damien, J.~Wakefield, and S.~Walker.
\newblock Gibbs sampling for bayesian non-conjugate and hierarchical models by
  using auxiliary variables.
\newblock {\em Journal of the Royal Statistical Society: Series B},
  61(2):331--344, 1999.

\bibitem{Miles:1964}
R.~E. Miles.
\newblock Random polygons determined by random lines in a plane.
\newblock {\em Proceedings of the National Academy of Science}, 52(4):901--907,
  1964.

\bibitem{Dunson:2006}
D.~B. Dunson.
\newblock Bayesian dynamic modeling of latent trait distributions.
\newblock {\em Biostatistics}, 7(4):511--568, 2006.

\bibitem{Ren:Dunson:Carin:2008}
L.~Ren, D.~B. Dunson, and L.~Carin.
\newblock The dynamic hierarchical {D}irichlet process.
\newblock In {\em International Conference on Machine Learning}, 2008.

\bibitem{Dunson:Pillai:Park:2007}
D.~B. Dunson, N.~Pillai, and J.-H. Park.
\newblock Bayesian density regression.
\newblock {\em JRSS:B}, 69(2):163--183, 2007.

\bibitem{Zhou:2012:BPFA}
M.~Zhou, H.~Chen, J.~W. Paisley, L.~Ren, L.~Li, Z.~Xing, D.~B. Dunson,
  G.~Sapiro, and L.~Carin.
\newblock Nonparametric {B}ayesian dictionary learning for analysis of noisy
  and incomplete images.
\newblock {\em IEEE Trans. Image Process.}, 21(1):130--144, 2012.

\bibitem{Ren:Dunson:Lindroth:Carin:2010}
L.~Ren, D.~B. Dunson, S.~Lindroth, and L.~Carin.
\newblock Dynamic nonparametric bayesian models for analysis of music.
\newblock {\em Journal of the American Statistical Association},
  105(490):458--472, 2010.

\bibitem{Blei:2003}
D.~M. Blei, A.~Y. Ng, and M.~I. Jordan.
\newblock Latent dirichlet allocation.
\newblock {\em Journal of Machine Learning Research}, 3:993--1022, 2003.

\bibitem{Ahmed:Xing:2010}
A.~Ahmed and E.~P. Xing.
\newblock Timeline: A dynamic hierarchical {D}irichlet process model for
  recovering birth/death and evolution of topics in text stream.
\newblock In {\em Uncertainty in Artificial Intelligence}, 2010.

\end{thebibliography}

\end{document}